\pdfoutput=1
\documentclass[11pt]{article}
\usepackage{authblk}

\usepackage[final]{acl}
\usepackage{times}
\usepackage{latexsym}
\usepackage[T1]{fontenc}
\usepackage{booktabs}
\usepackage{url}
\usepackage{float}
\usepackage[utf8]{inputenc}
\usepackage{microtype}
 \usepackage{graphicx}
\usepackage{authblk}

\usepackage{xcolor}

\def \thecolor {black}
\newcommand{\blue}[1]{\textcolor{\thecolor}{#1}}

\usepackage{array}
\makeatletter
\g@addto@macro{\endtabular}{\rowfont{}}% Clear row font
\makeatother
\newcommand{\rowfonttype}{}% Current row font
\newcommand{\rowfont}[1]{% Set current row font
\gdef\rowfonttype{#1}#1\ignorespaces%
}
\makeatother

\title{Probing neural language models\\ for understanding of words of estimative probability }

\author[1]{Damien Sileo}
\author[2]{Marie-Francine Moens}

\affil[1]{Univ. Lille, Inria, CNRS, Centrale Lille, UMR 9189 - CRIStAL, F-59000 Lille, France}
\affil[2]{Department of Computer Science, KU Leuven, Belgium} \vspace{-2ex}
\affil[ ]{\url{damien.sileo@inria.fr}}

\begin{document}
%\nolinenumbers 

\maketitle
\begin{abstract}
 %%%%%%%%%%%%%%%%%%%%%%%%%%%%%%%%%%%%%%%%%IMPORTANT%%%%%%%%%%%%%%%%%%%%%%%%%%
%\onecolumn
%Words of estimative probability (WEP) are expressions of a statement's plausibility (\textit{probably, maybe, likely, doubt, likely, unlikely, impossible}...). Multiple surveys demonstrate the agreement of human evaluators when assigning numerical probability levels to WEP. For example, \textit{highly likely} corresponds to a median chance of $0.90{\pm}0.08$ in \citet{fagen-ulmschneider}'s survey.
%In this work, we measure the ability of neural language processing models to capture the consensual probability level associated to each WEP. Firstly, we use the UNLI dataset \cite{chen-etal-2020-uncertain} which associates premises and hypotheses with their perceived joint  probability $p$, to construct prompts, e.g. "[\textsc{Premise}]. [\textsc{Wep}], [\textsc{Hypothesis}]." and assess whether language models can predict whether the WEP consensual probability level is close to $p$. Secondly, we construct a dataset of WEP-based probabilistic reasoning, to test whether language models can reason with WEP compositions. When prompted  "[\textsc{EventA}] \textit{is likely}. [\textsc{EventB}]  \textit{is impossible}.", a causal language model should not express that [\textsc{EventA$\&$B}] is likely. We show that both tasks are unsolved by off-the-shelf English language models, but that fine-tuning leads to transferable improvement. 

Words of Estimative Probability (WEP) are phrases used to express the plausibility of a statement. Examples include terms like \textit{probably, maybe, likely, doubt, unlikely}, and \textit{impossible}. Surveys have shown that human evaluators tend to agree when assigning numerical probability levels to these WEPs. For instance, the term \textit{highly likely} equates to a median probability of $0.90{\pm}0.08$ according to a survey by \citet{fagen-ulmschneider}.
In this study, our focus is to gauge the competency of neural language processing models in accurately capturing the consensual probability level associated with each WEP. Our first approach is utilizing the UNLI dataset \cite{chen-etal-2020-uncertain}, which links premises and hypotheses with their perceived joint probability $p$. From this, we craft prompts in the form: "[\textsc{Premise}]. [\textsc{Wep}], [\textsc{Hypothesis}]." This allows us to evaluate whether language models can predict if the consensual probability level of a WEP aligns closely with $p$.
In our second approach, we develop a dataset based on WEP-focused probabilistic reasoning to assess if language models can logically process WEP compositions. For example, given the prompt "[\textsc{EventA}] \textit{is likely}. [\textsc{EventB}] \textit{is impossible}.", a well-functioning language model should not conclude that [\textsc{EventA$\&$B}] is likely.
Through our study, we observe that both tasks present challenges to out-of-the-box English language models. However, we also demonstrate that fine-tuning these models can lead to significant and transferable improvements.

\end{abstract}

\section{Introduction}

Expression of uncertainty is an important part of communication.  Formal statistics are the rigorous way to quantify uncertainty but do not fit all communication styles. Words of estimative probability (WEP) such as \textit{maybe} and \textit{believe} are adverbs or verbs that are informal alternatives. \citet{kent1964words} noted the importance of clarifying WEP meaning for intelligence analysis in the Central Intelligence Agency, and provided guidelines for mapping WEP to numerical probabilities. Several studies then measured the human perceptions of probability words and discovered some agreement with \citet{kent1964words}'s guidelines. In this work, we use the scale derived from a survey \cite{fagen-ulmschneider}, which is the largest and most recent WEP perception survey available. 123 participants were asked to label WEP with numerical probabilities. We use the median of the participant answers to assign a consensual value to each WEP. Associated probabilities for the 19 WEP we use are available in Appendix \ref{sec:probs}, table \ref{tab:probs}.

Here, we assess whether neural language models learn the consensual probability judgment of WEP from language modeling pretraining. We develop datasets and a methodology to probe neural language model understanding of WEP. The first dataset leverages previously annotated probability scores between a premise and a hypothesis, in order to measure a language model's ability to capture the agreement between numerical probabilities and WEP-expressed probabilities. The second dataset is based on compositions of facts with WEP-expressed probabilities, and measures verbal probabilistic reasoning in language models.

Our contributions are as follows: (i) two datasets and methods to measure understanding of WEP; and  (ii) evaluation of the ability of neural language models (GPT2, RoBERTa-trained on MNLI) to tackle WEP-related problems, showing that off-the-shelf models are very little influenced by them, even though fine-tuning on our constructed datasets quickly leads to high accuracies. The code and generated datasets are publicly available\footnote{\href{https://huggingface.co/datasets/sileod/probability_words_nli}{\texttt{/hf.co/.../probability\_words\_nli}}}

\section{Related work}
\blue{Our work probes a particular aspect of language understanding. We do not analyze the inside of the models \citep{rogers-etal-2020-bertology}. We focus on the models' ability to perform controlled tasks \citep{naik-etal-2018-stress,richardson2020probing} involving WEP.} WEP were studied in the context of intelligence analysis and linguistics, our work is the first to look at them through natural language processing (NLP) models. Our study also pertains to NLP analyses of logical reasoning and probability problems, and to uncertainty in natural language inference tasks. 
\vspace{-0.1cm}
\paragraph{Linguistics study of WEP} \citet{kent1964words}'s seminal work was the first to link WEP and numerical probability estimates, with intelligence analysis motivations \citep{dhami2021words} and a prescriptivist approach. This inspired further quantifications of human perceptions of WEP, in the context of medical reports \citep{OBrien1989WordsON, ott2021words} and weather reports \citep{lenhardt2020likely}. \citet{fagen-ulmschneider} proposed  the largest survey up to date with 123 participants about general-domain WEP perception. 
\vspace{-0.1cm}

\paragraph{Logical and probabilistic reasoning} Another strand of work probes NLP text encoders capabilities, notably reasoning abilities. \citet{weston2015towards} probed understanding of specific problems like negation, spatial and temporal reasoning with the bAbI dataset. \citet{richardson2020probing} and \citet{han2022folio} probe understanding of first-order logic reasoning, \citet{sileo2023mindgames} probe epistemic logic reasoning. Our work is the first to address probabilistic logic, alongside \citet{ijcai2017-0556,suster-etal-2021-mapping} who construct a dataset of natural language probability problems, e.g., \textit{"A bag has 4 white and 8 blue marbles. You pull out one marble and it is blue. You pull out another marble, what is the probability of it being white?"}. They also rely on the ProbLog solver \citep{de2007problog}, but focus on numeric probability problems. By contrast, our work targets WEP, and textual probabilistic logical reasoning.

%\paragraph{Natural language inference and uncertainty}
\paragraph{Natural language inference, uncertainty, \blue{modality, evidentiality}}
Uncertainty was also studied in the context of natural language inference tasks. \citet{xzhou2022distnli} study the disagreement across annotators when labeling entailment relationships. \citet{ordinal-common-sense-inference} annotate graded entailment with 5 probability levels, and the UNLI dataset \citep{chen-etal-2020-uncertain} go further by annotating numerical probabilities.
\blue{
Our work also pertains to the study of modality \cite{modality, modalityannotation} and more particularly evidentiality \cite{su-etal-2010-evidentiality}, but where previous work focused on WEP.}

\begin{figure*}
 \centering
\includegraphics[width =0.99\textwidth]{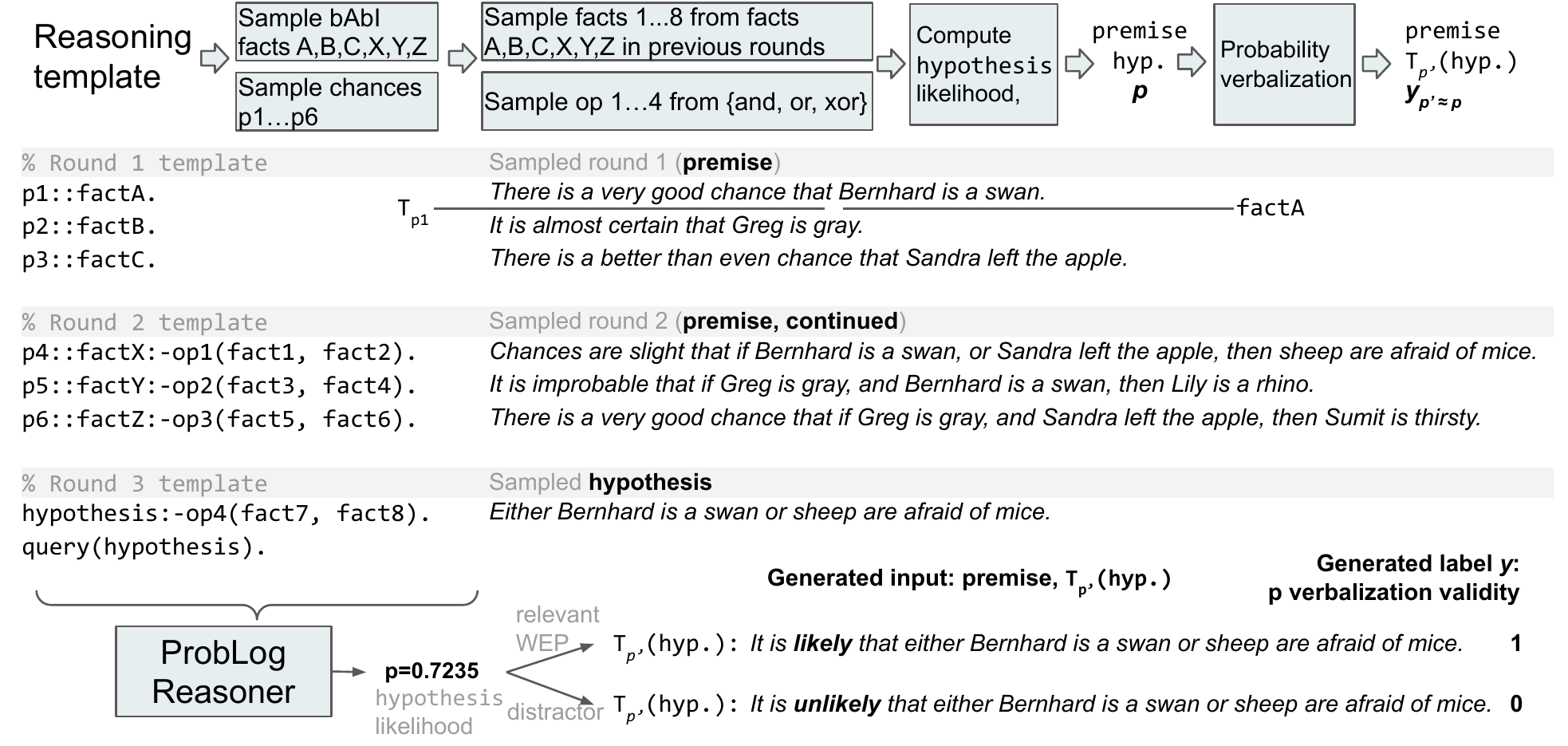}
  \caption{WEP-reasoning task constructions, with 2 hops.
  We sample randomly concrete facts $fact_i$ and probabilities $p_i$  then build modal sentences with verbalization templates. We randomly sample logical operators to compose the modal sentences from the previous rounds to construct a premise, then a hypothesis, and we use a probabilistic soft logic solver to compute the hypothesis probability. We then correctly and incorrectly verbalize this probability. This process generates data for the task of probability verbalization validity.  1 hop reasoning skips the second round: \texttt{fact7} and \texttt{fact8} are sampled from $\{$\texttt{factA},\texttt{factB},\texttt{factC}$\}$}
  
\label{fig:overview} 
\end{figure*}

\section{Probing WEP understanding}
\subsection{Verbalization and distractor generation \label{sec:verbalization}}

Our goal is to measure the understanding of WEP. One requirement of WEP understanding is capturing the consensual probability level. To test that, we use contexts (\textsc{Premise}) paired with a conclusions (\textsc{Hypothesis}). The likelihood of a conclusion, $p$, depends on the associated context. One example from UNLI \citep{chen-etal-2020-uncertain}, which annotates that, is (\textit{A man in a white shirt taking a picture
}, \textit{A man takes a picture }, 1.0). 

We convert a triplet (\textsc{Premise}, \textsc{Hypothesis}, $p$)
to the following verbalization:\vspace{-3mm}

\begin{equation}
    \textsc{Premise}.\;  T_p(\textsc{Hypothesis}). \vspace{-1mm}
\end{equation}

where  $T_p$ is a text template assigned to the probability $p$. To select a template, we find the WEP whose associated median probability (see table \ref{tab:probs}) is the closest to $p$. We then use handcrafted templates to construct a modal sentence from the selected WEP and the hypothesis, e.g., "\textit{It is  \textbf{certain} that \textbf{a man takes a picture}}". Table \ref{tab:templates} in appendix \ref{sec:appendix} displays the templates that we associate with each WEP. 

We also generate an invalid verbalization by randomly selecting an incorrect WEP (a WEP whose consensual probability differs from $p$ by at least $40\%$)\footnote{This threshold ensures sufficient distance, while also ensuring that each WEP has at least one possible distractor.}, e.g., \textit{It is unlikely that a man takes a picture}. We hypothesize that language models and entailment recognition models should give a higher score (respectively likelihood and entailment probability) to the correct valid verbalization than to the invalid verbalization of $p$.

\subsection{WEP-UNLI: probability/WEP matching}

The UNLI dataset annotates (\textsc{Premise}, \textsc{Hypothesis}) pairs from the SNLI dataset \citep{bowman-etal-2015-large} with joint probability scores $p$, totaling 55k training examples, 3k/3k validation/test examples. We use these examples to generate WEP-understanding dataset with verbalization validity prediction as shown in the previous subsection.

\subsection{WEP-Reasoning: WEP compositions}  
Here, our goal is to assess models' ability to reason over combinations of probabilistic statements. We construct synthetic (\textsc{Premise}, \textsc{Hypothesis}, $p$) examples from random \blue{factoids extracted from the bAbI dataset \citep{weston2015towards}}. Figure \ref{fig:overview} illustrates the construction of WEP-reasoning examples:

We randomly sample initial facts and associated probability levels, and we verbalize them with the previously mentioned templates from Table \ref{tab:templates} \blue{(Round 1)}.
We further compose them with randomly sampled logical operators (and, or, xor). We then generate a hypothesis with logical combinations of the previous round. Finally, we feed the constructed premise and hypothesis to a probabilistic soft reasoning engine in order to derive the likelihood of the hypothesis given the premise. We rely on the ProbLog \cite{de2007problog} reasoner which implements \citet{Dantsin} semantics.

\blue{To evaluate different complexities of reasoning, we propose two variants: \textbf{2-hop reasoning}, where facts in Round 2 combine facts from Round 1, and the final hypothesis combines facts from Round 2. and \textbf{1-hop reasoning} where facts from the hypothesis combine Round 1 facts (Round 2 is skipped).}

\begin{table*}
\centering
\small
\begin{tabular}{llll}
\toprule
                                 &    WEP-Reasoning (1 hop) &    WEP-Reasoning (2 hops) &             WEP-UNLI \\
\midrule
            Chance &          50.0 &          50.0 &          50.0 \\
\blue{Human baseline} &  \blue{97.0±1.0}& \blue{\textbf{93.5±1.5}}&  \blue{\textbf{89.5±2.5}}\\

\midrule
            GPT2 likelihood zero-shot &          50.1±0.0 &          50.0±0.0 &          45.6±0.0 \\
    RoBERTa likelihood zero-shot &          \blue{63.4±0.0} &          \blue{63.2±0.0} &          \blue{53.2±0.0} \\
RoBERTa-MNLI  zero-shot &          49.2±5.4 &          41.7±4.2 &          54.6±3.7 \\
\midrule
          RoBERTa+WEP-Reasoning (1 hop) fine-tuning& \textbf{97.8±0.4} &          81.6±1.3 &          61.2±0.4 \\
          RoBERTa+WEP-Reasoning (2 hops) fine-tuning &          85.0±1.6 & \textbf{91.1±0.1} &          62.3±1.7 \\
                   RoBERTa+WEP-UNLI fine-tuning &          62.4±0.4 &          64.3±0.1 & \textbf{84.4±0.5} \\
%\midrule
\bottomrule
\end{tabular}
\vspace{-1mm}
\caption{Test accuracy percentage of different models over the 3 WEP-understanding tasks. The last three rows display the accuracy when fine-tuning on each task, and transferability of the fine-tuned model outside the diagonal. \label{tab:downstream}\vspace{-3mm}}

\end{table*}

Since we want to sample more than two facts and we cannot a priori use text from the UNLI dataset, because UNLI only provides entailment likelihood for specific pairs. Combining several sentences could cause unaccounted interference. Therefore, we sample \blue{subject/verb/object} factoids from the bAbI \cite{weston2015towards} datasets instead, which is built with handwritten \blue{arbitrary factoids} such as \textit{John went to the kitchen}. To sample multiple factoids, we prevent any overlap of concepts \blue{(verb, subject, object)} between any pair of facts to make the facts independent of one another.

We sample probability levels from the list of medians of all WEP to prevent sampling the levels that too distant from a known WEP. When we assign a WEP to a probability level, we assume that the correct semantics is the consensual one, but humans differs slightly from this consensus. Still, when adding random perturbations of $20\%$ to sampled $p_{1...6}$, the hypothesis probability is perturbed by less than $40\%$ for $98\%$ of examples.

%To model this disagreement and its impact on the legitimacy of our probe, we added a random noise to the probability level before assigning a WEP. The disagreement across annotators has a limited impact on the labels of WEP-reasoning [TODO].

We generate 5k examples using the template depicted in Figure \ref{fig:overview}, and use $10\%/10\%$ of the data for the validation/test splits. \blue{Appendix \ref{sec:wepstats} shows the distribution of correct WEP for each dataset.}

\section{Experiments}
We conduct verbalization validity prediction (binary classification task of WEP correctness detection between two candidates) under two settings.
\vspace{-1mm}
\subsection{Zero-shot models}
We use off-the-shelf language models to assign likelihood scores to a context and its conclusion. We evaluate the rate at which valid verbalization is scored higher than invalid verbalization. \blue{
We refine the scores by also considering the average likelihood per token \cite{brown2020language,schick-schutze-2021-exploiting} and calibrated scores \cite{brown2020language,pmlr-v139-zhao21c} where we divide the score of a $\textsc{Premise}.\;  T_p(\textsc{Hypothesis}).$ by the score of $T_p(\textsc{Hypothesis}).$ We evaluate the normalized, length-normalized, and calibrated likelihood on the validation sets of each dataset and select the most accurate method for each dataset and model.}

We also consider a pretrained natural language inference model, which is trained to predict entailment scores between a context and a conclusion.

\paragraph{GPT2} We use the pretrained GPT2 base version with 127M parameters \cite{radford2019language}, which is a causal language model trained to estimate text likelihood. We concatenate the premise and hypothesis and compute their likelihood as a plausibility score.
\vspace{-1mm}
\paragraph{RoBERTa} We also use the pretrained RoBERTa base model with 123M parameters  \cite{liu2019roberta} to score the masked language modeling likelihood of the premise/hypothesis pair.
\vspace{-1mm}
\paragraph{RoBERTa-MNLI} We fine-tune RoBERTa on the MNLI entailment detection dataset \citep{williamsmnli18} with standard hyperparameters (see the following subsection).
\vspace{-1mm}
\paragraph{Human baseline} \blue{To establish human baseline performance on the constructed dataset, we had two NLP researchers annotate 100 examples randomly sampled from the test set of each dataset, with a multiple-choice question answering setting. Overall inter-annotator agreement is relatively high, with a Fleiss’s $\kappa$ of 0.70/0.68/0.71 for WEP Reasoning 1 hop, 2 hops and WEP-UNLI respectively.
}

\subsection{Fine-tuning and transfer across probes}

We fine-tune RoBERTa-base models on our datasets, using standard \citep{mosbach2021on} hyperparameters\footnote{Deviation from these hyperparameters did not yield significant improvement on the validation sets.} (3 epochs, sequence length of 256, learning rate of $2.10^{-5}$ batch size of 16. \blue{We use length-normalization with GPT2 likelihood and calibration with RoBERTa likelihood as they worked best on the validation sets.}).
We use a multiple-choice-question answering setup (we predict logit scores for the valid and invalid verbalization, combine their score with a softmax, then optimize the likelihood of the valid verbalization). The same format is applied to all tasks, so we can also study the transfer of capacities acquired during fine-tuning of each probe, for instance, between probability matching and compositional reasoning.

\subsection{Results and discussion}

Table \ref{tab:downstream} shows the results of our experiments. The very low accuracy of causal and masked language models (first two rows) demonstrates how challenging the WEP-understanding tasks are.

RoBERTa fine-tuned on MNLI dataset performs better than chance for WEP-UNLI. MNLI contains 814 instances of \textit{probably} in the MNLI dataset, but we found little to no evidence of WEP compositions among them, which can explain the results.

Finally, fine-tuning on the dataset of a particular probe leads to high test accuracy on the associated test set. More surprisingly, fine-tuning on one dataset also causes substantial accuracy gain \blue{on} other probes. This suggests that our datasets can be incorporated in text encoder training in order to improve WEP handling. 

\section{Conclusion}

We investigated WEP understanding in neural language models with new datasets and experiments, showing that WEP processing is challenging but helped by supervision which leads to transferable improvement. Future work could extract WEP probability scales from the UNLI dataset as an alternative to human perception surveys, but our work suggests that this requires language modeling progress.

\section{Acknowledgements}
This work is part of the CALCULUS project, which is
funded by the ERC Advanced Grant H2020-ERC-2017 
ADG 788506\footnote{\url{https://calculus-project.eu/}}. 

\nocite{srivastava2022beyond}

\bibliography{main}
\bibliographystyle{acl_natbib}

\clearpage
\onecolumn
\appendix

\section{Associated probabilities \label{sec:probs}}

\vspace{-0.1cm}
\begin{table}[H]
\begin{center}
\begin{tabular}{ll}
\toprule
WEP & \hspace*{-0.8cm}Median probability judgment\\
\midrule
\textit{certain}            &  $100^{\dag}$  \\
\textit{almost certain}     &  $95.0\pm10.9$  \\
\textit{highly likely}      &   $90.0\pm8.4$  \\
\textit{very good chance}   &  $80.0\pm10.8$  \\
\textit{we believe}         &  $75.0\pm15.0$  \\
\textit{likely}             &  $70.0\pm11.3$  \\
\textit{probably}           &  $70.0\pm12.9$  \\
\textit{probable}           &  $70.0\pm14.7$  \\
\textit{better than even}   &   $60.0\pm9.1$  \\
\textit{about even}         &   $50.0\pm4.9$  \\
\textit{probably not}       &  $25.0\pm14.4$  \\
\textit{we doubt}           &  $20.0\pm16.9$  \\
\textit{unlikely}           &  $20.0\pm15.0$  \\
\textit{little chance}      &  $10.0\pm12.2$  \\
\textit{chances are slight} &  $10.0\pm10.9$  \\
\textit{improbable}         &  $10.0\pm17.5$  \\
\textit{highly unlikely}    &   $5.0\pm17.3$  \\
\textit{almost no chance}   &   $2.0\pm17.0$  \\
\textit{impossible}         &    $0^{\dag}$  \\
\bottomrule
\end{tabular}
\end{center}
\caption{Median probability percentage associated to words of estimative probability according to \cite{fagen-ulmschneider}. First and last words ($\dag$) are taken from \cite{kent1964words}. \label{tab:probs}}
\end{table}

\section{WEP verbalization template\label{sec:appendix}}

\begin{table}[H]
\centering
\begin{tabular}{ll}
\toprule
                        WEP &                                                         Verbalization template \\
\midrule
        \textit{about even} &             \textit{chances are about even that} [\textsc{Fact}] \\
    \textit{almost certain} &               \textit{it is almost certain that} [\textsc{Fact}] \\
  \textit{almost no chance} &          \textit{there is almost no chance that} [\textsc{Fact}] \\
  \textit{better than even} & \textit{there is a better than even chance that} [\textsc{Fact}] \\
           \textit{certain} &                      \textit{it is certain that} [\textsc{Fact}] \\
\textit{chances are slight} &                 \textit{chances are slight that} [\textsc{Fact}] \\
     \textit{highly likely} &                \textit{it is highly likely that} [\textsc{Fact}] \\
   \textit{highly unlikely} &              \textit{it is highly unlikely that} [\textsc{Fact}] \\
        \textit{impossible} &                   \textit{it is impossible that} [\textsc{Fact}] \\
        \textit{improbable} &                   \textit{it is improbable that} [\textsc{Fact}] \\
            \textit{likely} &                       \textit{it is likely that} [\textsc{Fact}] \\
     \textit{little chance} &             \textit{there is little chance that} [\textsc{Fact}] \\
          \textit{probable} &                     \textit{it is probable that} [\textsc{Fact}] \\
          \textit{probably} &            \textit{it is probably the case that} [\textsc{Fact}] \\
      \textit{probably not} &        \textit{it is probably not the case that} [\textsc{Fact}] \\
          \textit{unlikely} &                     \textit{it is unlikely that} [\textsc{Fact}] \\
  \textit{very good chance} &        \textit{there is a very good chance that} [\textsc{Fact}] \\
        \textit{we believe} &                         \textit{we believe that} [\textsc{Fact}] \\
          \textit{we doubt} &                           \textit{we doubt that} [\textsc{Fact}] \\
\bottomrule
\end{tabular}
\caption{Templates used to convert a fact and a WEP expressed uncertainty into a modal sentence. \label{tab:templates}}
\end{table}

\begin{table}[H]

\end{table}

%\newpage

\section{\blue{WEP frequencies on the generated datasets} \label{sec:wepstats}}
\blue{
\begin{table}[H]
\centering
\small
\begin{tabular}{lrlrlr}
\toprule
WEP-reasoning&(1 hop)&WEP-Reasoning&(2 hops)&WEP-USNLI\\
\midrule
                      WEP &  frequency &                       WEP &  frequency &                       WEP &  frequency \\
\midrule
        \textit{about even} &              11.1 &         \textit{impossible} &              13.2 &         \textit{impossible} &              25.6 \\
      \textit{probably not} &               9.7 &         \textit{about even} &              10.8 &   \textit{better than even} &              10.7 \\
  \textit{better than even} &               7.7 &       \textit{probably not} &               9.0 &            \textit{certain} &               7.2 \\
        \textit{we believe} &               7.1 &    \textit{highly unlikely} &               8.2 &         \textit{about even} &               6.9 \\
     \textit{highly likely} &               6.4 &   \textit{almost no chance} &               8.0 &     \textit{almost certain} &               6.7 \\
           \textit{certain} &               6.0 &   \textit{better than even} &               6.6 &      \textit{highly likely} &               6.0 \\
   \textit{highly unlikely} &               5.9 &         \textit{we believe} &               4.3 &   \textit{very good chance} &               5.9 \\
  \textit{almost no chance} &               5.8 &      \textit{highly likely} &               4.0 &   \textit{almost no chance} &               5.0 \\
        \textit{impossible} &               5.3 &   \textit{very good chance} &               4.0 &         \textit{we believe} &               4.1 \\
    \textit{almost certain} &               5.1 &           \textit{we doubt} &               4.0 &    \textit{highly unlikely} &               4.1 \\
  \textit{very good chance} &               4.7 &         \textit{improbable} &               3.9 &       \textit{probably not} &               3.4 \\
\textit{chances are slight} &               3.6 & \textit{chances are slight} &               3.9 &             \textit{likely} &               2.5 \\
     \textit{little chance} &               3.5 &           \textit{unlikely} &               3.6 &           \textit{probable} &               2.4 \\
          \textit{probable} &               3.2 &      \textit{little chance} &               3.5 &           \textit{probably} &               2.4 \\
          \textit{unlikely} &               3.1 &     \textit{almost certain} &               2.9 &           \textit{unlikely} &               1.5 \\
            \textit{likely} &               3.1 &            \textit{certain} &               2.7 &      \textit{little chance} &               1.5 \\
          \textit{probably} &               3.0 &             \textit{likely} &               2.5 & \textit{chances are slight} &               1.5 \\
          \textit{we doubt} &               2.9 &           \textit{probable} &               2.4 &         \textit{improbable} &               1.4 \\
        \textit{improbable} &               2.9 &           \textit{probably} &               2.2 &           \textit{we doubt} &               1.4 \\
\bottomrule
\end{tabular}
\caption{\blue{Validation set frequency of WEP in the correct answer of each dataset (percentages).}}
\end{table}
}
\end{document}